\documentclass[conference]{IEEEtran}
\IEEEoverridecommandlockouts
\usepackage{cite}
\usepackage{amsmath,amssymb,amsfonts}
\usepackage{graphicx}
\usepackage{textcomp}
\usepackage{xcolor}
\def\BibTeX{{\rm B\kern-.05em{\sc i\kern-.025em b}\kern-.08em
    T\kern-.1667em\lower.7ex\hbox{E}\kern-.125emX}}

\usepackage{algorithm} 
\usepackage{algpseudocode}
\usepackage[algo2e,linesnumbered]{algorithm2e} 

\usepackage{scalerel}
\usepackage{tikz}
\usetikzlibrary{svg.path}

\definecolor{orcidlogocol}{HTML}{A6CE39}
\tikzset{
    orcidlogo/.pic={
        \fill[orcidlogocol] svg{M256,128c0,70.7-57.3,128-128,128C57.3,256,0,198.7,0,128C0,57.3,57.3,0,128,0C198.7,0,256,57.3,256,128z};
        \fill[white] svg{M86.3,186.2H70.9V79.1h15.4v48.4V186.2z}
        svg{M108.9,79.1h41.6c39.6,0,57,28.3,57,53.6c0,27.5-21.5,53.6-56.8,53.6h-41.8V79.1z M124.3,172.4h24.5c34.9,0,42.9-26.5,42.9-39.7c0-21.5-13.7-39.7-43.7-39.7h-23.7V172.4z}
        svg{M88.7,56.8c0,5.5-4.5,10.1-10.1,10.1c-5.6,0-10.1-4.6-10.1-10.1c0-5.6,4.5-10.1,10.1-10.1C84.2,46.7,88.7,51.3,88.7,56.8z};
    }
}

\newcommand\orcidicon[1]{\href{https://orcid.org/#1}{\mbox{\scalerel*{
                \begin{tikzpicture}[yscale=-1,transform shape]
                \pic{orcidlogo};
                \end{tikzpicture}
            }{|}}}}

\usepackage[pagebackref=true,breaklinks=true,colorlinks,citecolor=blue]{hyperref}
\usepackage{balance}
\usepackage{tabularx,booktabs}
\usepackage{array, multirow, boldline}
\usepackage{subcaption}

\begin{document}

\title{Low-Light Enhancement via Encoder-Decoder Network with Illumination Guidance}

\author{
\IEEEauthorblockN{Le-Anh Tran$^{\textsuperscript{\orcidicon{0000-0002-9380-7166}}}$,
Chung Nguyen Tran$^{\textsuperscript{\orcidicon{0009-0000-6402-8799}}}$,
Ngoc-Luu Nguyen,
Nhan Cach Dang$^{\textsuperscript{\orcidicon{0000-0001-6979-9197}}}$, \\
Jordi Carrabina$^{\textsuperscript{\orcidicon{0000-0002-9540-8759}}}$,
David Castells-Rufas$^{\textsuperscript{\orcidicon{0000-0002-7181-9705}}}$, and
Minh Son Nguyen$^{\textsuperscript{\orcidicon{0000-0001-5513-3992}}}$
}
}

\maketitle

\begin{abstract}

This paper introduces a novel deep learning framework for low-light image enhancement, named the Encoder-Decoder Network with Illumination Guidance (EDNIG). Building upon the U-Net architecture, EDNIG integrates an illumination map, derived from Bright Channel Prior (BCP), as a guidance input. This illumination guidance helps the network focus on underexposed regions, effectively steering the enhancement process. To further improve the model's representational power, a Spatial Pyramid Pooling (SPP) module is incorporated to extract multi-scale contextual features, enabling better handling of diverse lighting conditions. Additionally, the Swish activation function is employed to ensure smoother gradient propagation during training. EDNIG is optimized within a Generative Adversarial Network (GAN) framework using a composite loss function that combines adversarial loss, pixel-wise mean squared error (MSE), and perceptual loss. Experimental results show that EDNIG achieves competitive performance compared to state-of-the-art methods in quantitative metrics and visual quality, while maintaining lower model complexity, demonstrating its suitability for real-world applications. The source code for this work is available at \href{https://github.com/tranleanh/ednig}{https://github.com/tranleanh/ednig}.

\end{abstract}



\section{Introduction}

\begin{figure*}[t]
\centering
\includegraphics[width=1.0\textwidth]{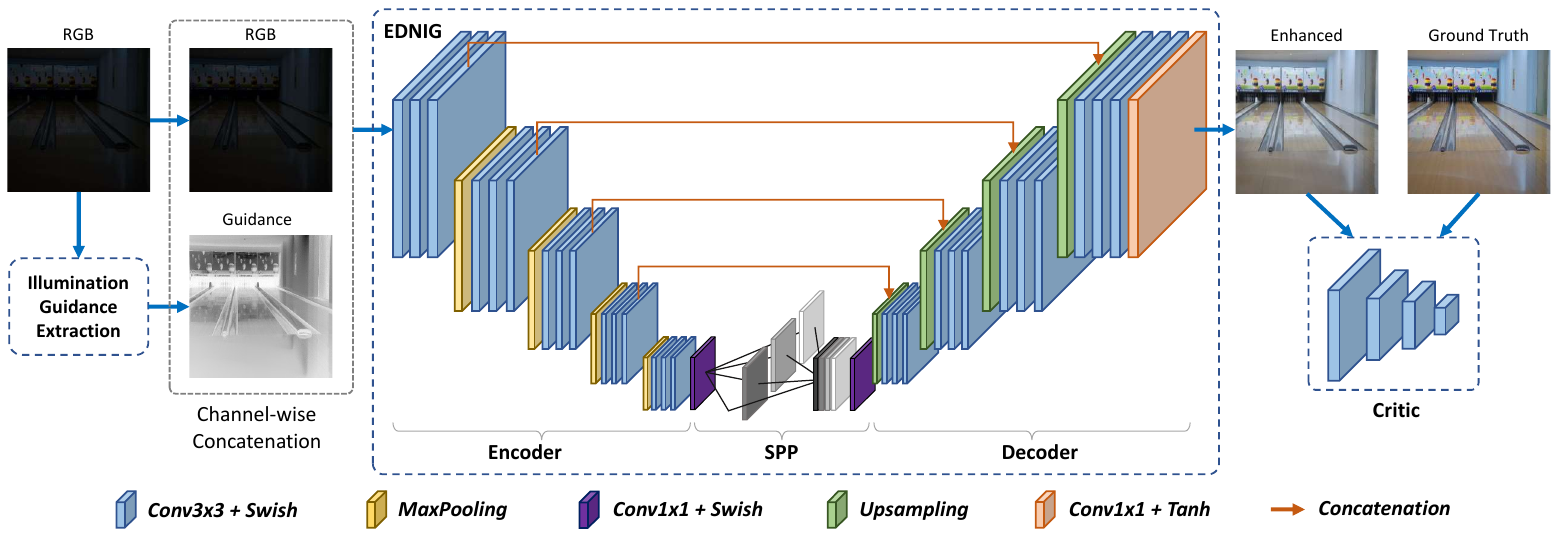}
\caption{The EDNIG framework.}
\label{fig:ednig}
\end{figure*}

Low-light image enhancement remains a critical challenge in computer vision, with applications spanning night-time photography, surveillance, autonomous driving, and medical imaging. Images captured under low-light conditions often suffer from poor visibility, reduced contrast, and amplified noise, which degrade both human perception and the performance of downstream tasks such as object detection and recognition. Traditional approaches to this problem, including histogram equalization \cite{jebadass2022low} and gamma correction \cite{jeong2021optimization}, tend to produce limited improvements, often introducing artifacts or failing to recover fine details. 

In recent years, deep learning techniques, such as convolutional neural networks (CNNs) and generative adversarial networks (GANs) \cite{goodfellow2014generative}, have emerged as powerful tools for image processing, leveraging large-scale data and sophisticated encoder-decoder architectures for image-to-image translation tasks to achieve impressive results \cite{ronneberger2015u,tran2019robust,tran2024lightweight}. A notable example is the Encoder-Decoder Network with Guided Transmission Map (EDN-GTM) \cite{tran2022novel,tran2022encoder,tran2024encoder}, originally proposed for single-image dehazing. EDN-GTM integrates a transmission map derived from Dark Channel Prior (DCP) \cite{he2010single} into a U-Net framework \cite{ronneberger2015u}, guiding the network to perform haze removal effectively. Although EDN-GTM performs exceptionally well in haze removal, its dependence on transmission maps, which are specifically designed for atmospheric scattering, restricts its direct use in low-light enhancement, where illumination distribution is a more critical factor. Nevertheless, the overall architecture of EDN-GTM remains promising for image restoration tasks thanks to its strengths, including multi-scale feature extraction with Spatial Pyramid Pooling (SPP) \cite{tran2024encoder}.

Inspired by the success of EDN-GTM in image restoration, we propose a novel adaptation of the EDN-GTM framework for low-light image enhancement. Specifically, we replace the transmission map with an illumination map derived from Bright Channel Prior (BCP) \cite{lee2020unsupervised}, and introduce the Encoder-Decoder Network with Illumination Guidance (ENDIG). The illumination map, which estimates the spatial distribution of light intensity in a scene, serves as a natural guide for this task. By feeding both the low-light RGB image and its illumination map into an encoder-decoder network, ENDIG learns to enhance visibility while preserving structural integrity and suppressing noise. By shifting the focus from transmission to illumination, we aim to not only advance low-light image enhancement but also provide a flexible framework adaptable to related image restoration tasks. In a nutshell, the contributions of this work are threefold: 1) an innovative application of illumination maps to guide a deep learning pipeline, 2) a refined network architecture tailored for low-light restoration, and 3) an experimental evaluation across diverse scenarios, demonstrating enhanced performance in both visual quality and quantitative metrics with reduced model complexity.

The remainder of this paper is organized as follows: Section \ref{sec:relatedwork} briefly describes the related work. The proposed framework is elaborated in Section \ref{sec:methodology}. The experimental results are presented in Section \ref{sec:experiments}. Section \ref{sec:conclusions} concludes the paper.

\section{Related Work}
\label{sec:relatedwork}

Early low-light image enhancement methods relied on traditional image processing techniques, such as histogram equalization and its variants, which redistribute pixel intensities to enhance contrast \cite{jebadass2022low,reza2004realization}. While computationally efficient, these approaches often amplify noise and fail to recover details in severely underexposed regions. Retinex-based methods, grounded in the theory of human color perception, decompose images into reflectance and illumination components to adjust lighting \cite{land1977retinex}. Single-scale Retinex (SSR) \cite{jobson1997properties} and multi-scale Retinex (MSR) \cite{rahman1996multi} have shown promise, but their performance is sensitive to parameter tuning and struggles with complex scenes. Optimization-based techniques, such as LIME \cite{guo2016lime}, estimate illumination maps using hand-crafted priors and refine them via regularization. Although effective in some cases, these methods lack the generalization ability required for diverse low-light conditions.

Currently, deep learning has transformed low-light image enhancement by enabling data-driven feature learning, with encoder-decoder architectures such as U-Net \cite{ronneberger2015u} demonstrating exceptional effectiveness. Retinex-Net \cite{wei2018deep} integrates the Retinex theory \cite{land1977retinex} with deep learning via training separate networks for decomposition and illumination adjustment. GLADNet \cite{wang2018gladnet} leverages the global priori knowledge of illumination for detail reconstruction, while KinD \cite{zhang2019kindling} decomposes images into reflectance and illumination, using dedicated networks for detail restoration and brightness adjustment. KinD++ \cite{zhang2021beyond} enhances KinD with improved illumination estimation and noise suppression. Uformer \cite{wang2022uformer} introduces a transformer-based approach for better detail recovery. WeatherDiff \cite{ozdenizci2023restoring} addresses low-light enhancement under adverse weather conditions using a size-agnostic denoising process with smoothed noise estimates. PQP \cite{wang2024zero} employs an illumination-invariant prior, guided by physical light transfer theory, to restore images via a trainable prior-to-image framework. These methods achieve realistic enhancement and structural fidelity, highlighting the promise of neural networks in this domain.

Guided deep learning models, which incorporate domain-specific priors into neural networks, offer a compelling middle ground between traditional and fully data-driven approaches. In the context of image restoration, EDN-GTM \cite{tran2024encoder} exemplifies this paradigm by using a transmission map, estimated via DCP \cite{he2010single}, to guide a U-Net-based network for image dehazing, achieving impressive performance. The auxiliary transmission map provides depth-related information, enabling the network to adaptively handle varying haze densities. Building on the success of EDN-GTM, the proposed framework replaces the transmission map with an illumination map tailored to light distribution, addressing the unique challenges of low-light scenes. While prior illumination-guided models exist, the proposed EDNIG approach distinguishes itself by leveraging different, straightforward, and effective architectural components and a hybrid loss function, drawing inspiration from EDN-GTM's design. This positions EDNIG as a novel contribution at the intersection of guided deep learning and low-light enhancement, aiming to deliver both robustness and high-quality restoration.

\begin{figure*}[t]
\centering
\includegraphics[width=1.0\textwidth]{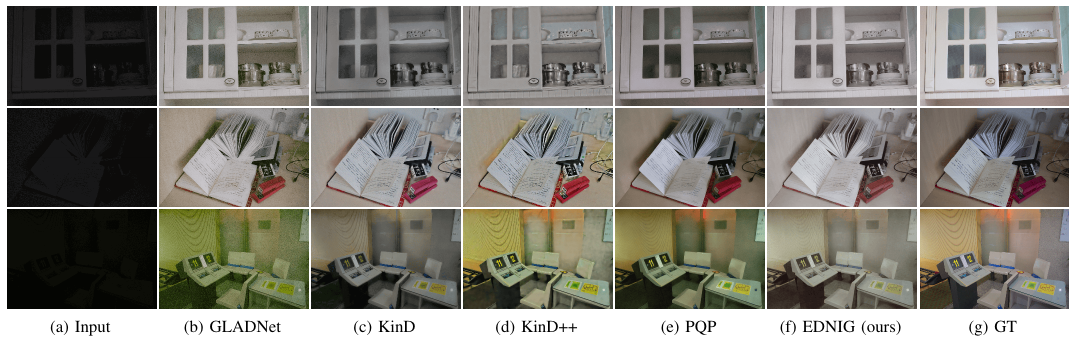}
\caption{Typical visual comparisons of various approaches on the LOL validation dataset.}
\label{fig:loldataset}
\end{figure*}

\section{Methodology}
\label{sec:methodology}

\subsection{Illumination Guidance Extraction}
\label{subsec:illumination}

To extract the illumination guidance map $M$, we adopt a statistical prior, BCP \cite{lee2020unsupervised}, which assumes that in natural, well-lit images, at least one RGB channel in a local patch has high intensity. Specifically, for an input low-light image $I$, we first estimate an initial illumination map $\tilde{M}$ by taking the maximum intensity across \{$R$, $G$, $B$\} channels at each local patch:
\begin{equation}
    \tilde{M}(x) = \max_{y \in \alpha(x)} \max_{c \in \{R,G,B\}} I_c(y),
\end{equation}
where $I_c(.)$ is the intensity of channel $c$ at pixel $y$ in local patch $\alpha(.)$ centered at $x$. To refine $\tilde{M}$ and reduce noise or inconsistencies, guided filtering is applied \cite{he2012guided}, producing a smoothed illumination map $M$. The guided filter preserves edges aligned with the input image while suppressing artifacts, making $M$ a robust guidance for the network. The resulting $M$ is concatenated with $I$, forming a four-channel input \{$R$, $G$, $B$, $M$\} to the network.

\subsection{Network Architecture}
\label{subsec:architecture}

Fig. \ref{fig:ednig} depicts the structure of the proposed framework. The generator (EDNIG) architecture is derived from the EDN-GTM design, modified specifically for low-light enhancement. To boost efficiency, we reduce the number of channels in the first layer to 12, down from 64 in the original EDN-GTM setup, and double them with each downscaling step. This results in a significantly compact model with only 1.74M parameters compared to the 33.26M parameters of EDN-GTM. As shown in Fig. \ref{fig:ednig}, EDNIG consists of an encoder-decoder structure and is trained within a GAN framework \cite{goodfellow2014generative}. The encoder extracts features from the four-channel input through five convolution stages. Each stage comprises three $Conv 3\times3$ layers with stride $1$ and Swish activation. The Swish activation function is chosen over ReLU and LeakyReLU due to its smooth non-linearity, which enhances gradient flow and captures complex patterns. Max pooling (\( 2\times2 \), stride 2) is applied 4 times, reducing spatial dimensions, yielding feature maps at scales 1/2, 1/4, 1/8, and 1/16 of the input resolution. 
At the bottleneck, the SPP module applies pooling operations with kernel sizes of \( 5\times5 \), \( 9\times9 \), and \( 13\times13 \), capturing multi-scale contextual information. Specifically, small size ($5\times5$) captures fine-grained details, medium size ($9\times9$) balances local and global context, and large size ($13\times13$) aggregates broader contextual information, improving the model's capability to process features across diverse scales, thereby boosting overall performance.
The decoder then upscales the bottleneck features back to the input resolution through four levels, each including a concatenation with corresponding encoded features. The final layer uses a $Conv 1\times1$ layer with Tanh activation to output an enhanced image. On the other hand, the critic retains the same configuration as in EDN-GTM \cite{tran2024encoder}, but applies a similar approach to channel reduction as in the generator.

\subsection{Loss Functions}
\label{subsec:loss}

To train the generator (EDNIG), an integral loss function is employed, combining adversarial realism, mean squared error (MSE), and perceptual quality. The adversarial loss $\mathcal{L}_{Adv}$ is defined as \cite{kupyn2018deblurgan}: 
\begin{equation}
\mathcal{L}_{Adv} = \frac{1}{B} \sum_{i=1}^{B} -C(G(I)),
\end{equation} where $B$, $C(.)$, $G(.)$, and $I$ denote the batch size, critic's output, generated image, and low-light input image, respectively. The MSE loss $\mathcal{L}_{MSE}$ is expressed as: 
\begin{equation}
\mathcal{L}_{MSE} = \frac{1}{N} \sum_{i=1}^{N} (G_i(z)-Y_i)^2,
\end{equation} 
where $N$ indicates the number of pixels in the input image, while $Y$ represents the ground truth. The perceptual loss $\mathcal{L}_{Per}$ to optimize the perceptual similarity is defined as \cite{johnson2016perceptual}: 
\begin{equation}
\mathcal{L}_{Per} = \frac{1}{M} \sum_{i=1}^{M} (\phi_i(G(I)) - \phi_i(Y))^2,
\end{equation} where $M$ denotes the number of elements in the feature map $\phi$ after the VGG16 model's $Conv3-3$ layer, and $\phi_i$  represents the $i^{th}$ activated value of $\phi$.

By combining all the related loss functions, the integral loss function $\mathcal{L}_{Gen}$ for optimizing EDNIG is formulated as: \begin{equation}
\mathcal{L}_{Gen} = \lambda_1 \mathcal{L}_{Adv} + \lambda_2 \mathcal{L}_{MSE} + \lambda_3 \mathcal{L}_{Per},
\end{equation} 
while the loss function $\mathcal{L}_{Cri}$ for training the critic is defined as below: 
\begin{equation}
\mathcal{L}_{Cri} = \lambda_4 \frac{1}{B} \sum_{i=1}^{B} (C(Y) - C(G(I))).
\end{equation}
Following \cite{tran2022encoder}, the balancing weight values are set as $\lambda_1$ = 100, $\lambda_2$ = 100, $\lambda_3$ = 100, and $\lambda_4$ = 1.

\begin{figure*}[t]
\centering
\includegraphics[width=1.0\textwidth]{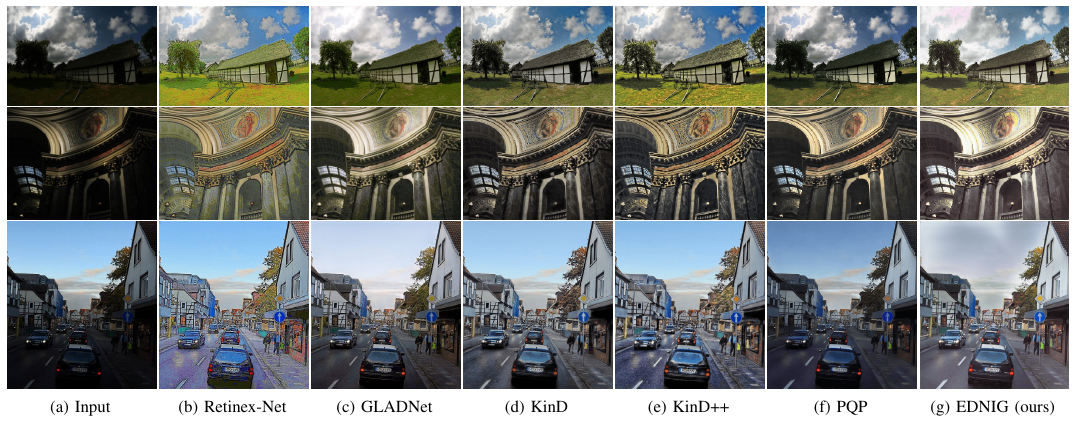}
\caption{Typical visual comparisons of various approaches on realistic image data.}
\label{fig:realistic}
\end{figure*}

\section{Experiments}
\label{sec:experiments}

\begin{table}[t]
\centering
\caption{Quantitative comparisons on the LOL dataset. The \textbf{bold} and \underline{underline} values denote the \textbf{best} and \underline{second-best} results, respectively.}
\begin{tabular}{llccr}
\toprule
Method & Year & PSNR$\uparrow$ & SSIM$\uparrow$ & \#Param(M) \\
\midrule
Retinex-Net \cite{wei2018deep} & 2018 & 16.774 & 0.5594 & 0.838  \\
GLADNet \cite{wang2018gladnet} & 2018 & 19.718 & 0.7035 & 1.128 \\
KinD \cite{zhang2019kindling} & 2019 & 20.726 & 0.8103 & 8.540 \\
Zero-DCE \cite{guo2020zero} & 2020 & 14.870 & 0.6600 & 0.079 \\
Kind++ \cite{zhang2021beyond} & 2021 & 21.300 & 0.8226 & 8.275 \\
DRBN \cite{yang2021band} & 2021	& 20.130 & \underline{0.8300} & 5.270 \\
Uformer \cite{wang2022uformer} & 2022 & 18.547 & 0.7212 & 5.290 \\
Restomer \cite{zamir2022restormer} & 2022 & 20.410 & 0.8060 & 26.130 \\
WeatherDiff \cite{ozdenizci2023restoring} & 2023 & 17.913 & 0.8110 & 82.960 \\
IR-SDE \cite{luo2023image} & 2023 & 20.450 & 0.7870 & 34.200 \\
IDR \cite{zhang2023ingredient}	& 2023 & \underline{21.340} & 0.8260 & 15.340 \\
PQP \cite{wang2024zero} & 2024 & 20.310 & 0.8080 & 0.327 \\
\midrule
EDNIG (ours) & & \textbf{21.512} & \textbf{0.8313} & 1.737 \\
\bottomrule
\end{tabular}
\label{table:results_on_lol}
\end{table}

\subsection{Experimental Settings}

The experiments were carried out on a Linux environment with GeForceGTX TITAN X GPUs and Intel(R) Xeon(R) Gold 6134 @3.20 GHz CPU. The proposed framework is implemented using the Tensorflow library. The dataset used is the LOL (LOw-Light) dataset \cite{wei2018deep}, which provides 500 low/normal-light image pairs, with 485 and 15 image pairs for training and validation, respectively. The network is optimized using Adam optimizer \cite{kingma2014adam} with a batch size of 1 and a learning rate of $10^{-4}$, decaying gradually to zero over 200 epochs. The input size of the network is $512\times512$. Data augmentation methods, including random flip and random crop, are employed to enhance the learning process. Quantitative performance on paired data is measured using PSNR and SSIM, whereas that on natural data is evaluated using NIQE \cite{mittal2012making} and BRISQUE \cite{mittal2012no}. Higher PSNR and SSIM values indicate superior performance, whereas lower NIQE and BRISQUE scores reflect better results. On the other hand, the computational complexity is assessed based on the number of parameters (\#Params).

\begin{table}[t]
\centering
\caption{Quantitative comparisons on realistic image data. The \textbf{bold} and \underline{underline} values denote the \textbf{best} and \underline{second-best} results, respectively.}
\begin{tabular}{llccr}
\toprule
Method & Year & NIQE$\downarrow$ & BRISQUE$\downarrow$ & \#Param(M) \\
\midrule
Retinex-Net \cite{wei2018deep} & 2018 & 15.731 & 21.994 & 0.838  \\
GLADNet \cite{wang2018gladnet} & 2018 & 13.734 & 23.372 & 1.128 \\
KinD \cite{zhang2019kindling} & 2019 & 15.041 & 24.738 & 8.540 \\
Kind++ \cite{zhang2021beyond} & 2021 & 14.265 & 21.246 & 8.275 \\
PQP \cite{wang2024zero} & 2024 & \textbf{13.253} & \textbf{20.309} & 0.327 \\
\midrule
EDNIG (ours) & & \underline{13.443} & \underline{20.515} & 1.737 \\
\bottomrule
\end{tabular}
\label{table:results_on_realistic}
\end{table}

\subsection{Quantitative Analysis}

The quantitative performance of the proposed EDNIG model is evaluated on the LOL dataset and compared against that of various state-of-the-art approaches. As summarized in Table \ref{table:results_on_lol}, EDNIG achieves a PSNR of 21.512 and an SSIM of 0.8313, surpassing all competing methods in both metrics. Compared to earlier methods like Retinex-Net, GLADNet, and KinD, EDNIG exhibits substantial improvements. More recent methods, including KinD++ and DRBN, are also outperformed by EDNIG. Furthermore, when compared to advanced techniques such as IR-SDE, Restomer, and IDR, EDNIG consistently delivers higher PSNR and SSIM scores, confirming its superior performance.

In terms of model complexity, EDNIG maintains a lightweight architecture with 1.737M parameters, which is comparable to methods like Retinex-Net (0.838M) and GLADNet (1.128M), and significantly more compact as compared to recently advanced models such as WeatherDiff (82.96M) and IR-SDE (34.20M). This balance of high performance and low computational complexity makes EDNIG particularly suitable for practical applications where resource constraints are a key consideration.

In addition, Table \ref{table:results_on_realistic} presents a quantitative performance comparison using a set of 10 standard realistic test images. As reported in Table \ref{table:results_on_realistic}, PQP and EDNIG deliver the top results. PQP achieves the best scores with an NIQE of 13.253 and a BRISQUE of 20.309, while EDNIG follows closely with an NIQE of 13.443 and a BRISQUE of 20.515, securing second place in both metrics.

Overall, the quantitative results on the LOL dataset and realistic data demonstrate that EDNIG achieves a strong balance of image quality and efficiency, outperforming most state-of-the-art methods in terms of quantitative measures while maintaining a relatively lightweight model size.

\subsection{Qualitative Analysis}

The qualitative comparisons on the LOL validation dataset and realistic image data are presented in Fig. \ref{fig:loldataset} and Fig. \ref{fig:realistic}, respectively. The comparisons involve several state-of-the-art approaches, including GLADNet, KinD, KinD++, PQP, Retinex-Net, and the proposed EDNIG method. The ground truth (GT) is provided for the LOL dataset to serve as a reference for ideal restoration.

As shown in Fig. \ref{fig:loldataset}, GLADNet and KinD introduce some degree of brightness enhancement but often struggle with color distortion and noise amplification. KinD++ improves upon KinD with better structural preservation, while PQP enhances contrast but introduces unnatural tone shifts. EDNIG, on the other hand, effectively recovers both illumination and structural details while preserving color consistency and reducing unwanted artifacts and noise. 

Fig. \ref{fig:realistic} presents qualitative comparisons on typical real-world image data. These images lack ground truth references, making it essential to assess perceptual quality subjectively. As illustrated in Fig. \ref{fig:realistic}, Retinex-Net introduces significant color distortions and unnatural artifacts, while GLADNet produces inconsistent brightness adjustments. KinD and KinD++ improve overall visibility but tend to amplify saturation and edge artifacts. PQP provides competitive results but occasionally alters the natural scene appearance. In contrast, EDNIG consistently achieves visually appealing enhancements, maintaining realistic color reproduction, preserving fine details, and avoiding excessive overexposure. 

Overall, the qualitative analysis demonstrates that EDNIG is competitive against existing methods in low-light image enhancement by achieving a superior balance between brightness correction, noise suppression, and color fidelity. These results highlight the potential of EDNIG for practical applications in real-world environments.

\begin{table}
  \caption{Average runtime (in seconds) of various methods. The \textbf{bold} and \underline{underline} values denote the \textbf{best} and \underline{second-best} results, respectively.}
  \centering
  \begin{tabular}{lcc}
    \toprule
    Method & Framework & Runtime (sec.) \\
    \midrule
    Retinex-Net \cite{wei2018deep} & TensorFlow &  0.121 \\
    GLADNet \cite{wang2018gladnet} & TensorFlow & \textbf{0.035} \\
    KinD \cite{zhang2019kindling} & TensorFlow & 0.192 \\
    Kind++ \cite{zhang2021beyond} & TensorFlow & 0.189 \\
    PQP \cite{wang2024zero} & PyTorch & 3.254 \\
    \midrule
    EDNIG (ours) & TensorFlow & \underline{0.042} \\
    \bottomrule
  \end{tabular}
  \label{table:runtime}
\end{table}

\subsection{Runtime Analysis}

While \#Params offers some insight into the model complexity to a certain extent, it is not comprehensive enough to consider several inference-related factors such as memory access, degree of parallelism, and platform characteristics \cite{tran2025distilled}. Hence, we have conducted a runtime comparison of several modern approaches. 

Table \ref{table:runtime} compares the average runtime (in seconds) of different methods on the LOL validation set. As summarized in Table \ref{table:runtime}, GLADNet achieves the fastest runtime at 0.035 seconds, followed by the proposed EDNIG at 0.042 seconds, both built based on TensorFlow. Retinex-Net, KinD, and KinD++ record runtimes of 0.121, 0.192, and 0.189 seconds, respectively, also on TensorFlow, indicating moderate efficiency. PQP, implemented in PyTorch, is the slowest at 3.254 seconds, likely due to the framework's architectural complexity. The results highlight GLADNet and EDNIG as the most efficient, with EDNIG offering a competitive runtime while maintaining strong quality metrics.

\section{Conclusions}
\label{sec:conclusions}

In this paper, we present EDNIG, a deep learning framework for low-light image enhancement that enhances the U-Net architecture by integrating an illumination map derived from BCP. This design guides the model to focus more effectively on underexposed regions, while SPP module enables robust multi-scale feature extraction to manage diverse lighting conditions. Optimized within a GAN framework using a composite loss function, comprising adversarial, mean squared error, and perceptual losses, EDNIG consistently produces competitive results compared to state-of-the-art approaches in both quantitative metrics and visual quality. Evaluations on a benchmark dataset and realistic low-light images highlight the proposed EDNIG model's strong generalization and practical utility. Future work will focus on improving illumination estimation, integrating attention mechanisms, and extending the framework to real-time video enhancement and more diverse datasets to further advance low-light image processing.

\balance
\bibliographystyle{IEEEtran}
\bibliography{ref.bib}

\end{document}